\documentclass[a4paper, 10pt, conference]{ieeeconf}      
\usepackage{FG2024}
\usepackage{times}
\usepackage{epsfig}
\usepackage{graphicx}
\usepackage{amsmath}
\usepackage{amssymb}
\usepackage{algorithmic}
\usepackage{graphicx}
\usepackage{textcomp}
\usepackage{xcolor}
\usepackage{longtable}
\usepackage{subcaption}
 \usepackage{dblfloatfix}
\usepackage{float}
 \usepackage{booktabs}
\usepackage{bigstrut}
\usepackage{multirow}
\usepackage{color}
\usepackage{caption}
\usepackage{fancyhdr}
\usepackage{listings} 
\usepackage{changepage}
\usepackage{pgfplots}
\usepackage{balance}
\let\labelindent\relax
\usepackage{enumitem}
\usepackage{fancyhdr}

\FGfinalcopy 

\IEEEoverridecommandlockouts                              
\def\FGPaperID{****} 

\title{\LARGE \bf
 3D Face Morphing Attack Generation using Non-Rigid Registration}

\author{\parbox{16cm}{\centering
    {\large Jag Mohan Singh and Raghavendra Ramachandra}\\
    {\normalsize
   Norwegian University of Science and Technology (NTNU), Norway}}
    \thanks{}
}

\pgfplotsset{compat=1.18}
\begin{document}

\maketitle
\thispagestyle{fancy}
\renewcommand{\headrulewidth}{0pt}
\fancyhf{}
\fancyhead[C]{2024 18th International Conference on Automatic Face and Gesture Recognition (FG)}





\fancyfoot[L]{979-8-3503-9494-8/24/\$31.00 \copyright 2024 IEEE}

\begin{abstract}

Face Recognition Systems (FRS) are widely used in commercial environments, such as e-commerce and e-banking, owing to their high accuracy in real-world conditions. However, these systems are vulnerable to facial morphing attacks, which are generated by blending face color images of different subjects. This paper presents a new method for generating 3D face morphs from two bona fide point clouds. The proposed method first selects bona fide point clouds with neutral expressions. The two input point clouds were then registered using a Bayesian Coherent Point Drift (BCPD) without optimization, and the geometry and color of the registered point clouds were averaged to generate a face morphing point cloud. The proposed method generates 388 face-morphing point clouds from 200 bona fide subjects. The effectiveness of the method was demonstrated through extensive vulnerability experiments, achieving a Generalized Morphing Attack Potential (G-MAP) of 97.93\%, which is superior to the existing state-of-the-art (SOTA) with a G-MAP of 81.61\%.
\end{abstract}

\section{Introduction \& Related Work}\label{Sec:intro}
 Face Recognition Systems (FRS) have achieved high levels of accuracy in uncontrolled, real-world environments, largely owing to advances in deep learning algorithms, as documented in the literature~\cite{schroff2015facenet,deng2019arcface}. This high level of accuracy has led to the adoption of FRS in various commercial settings including e-commerce and e-banking. In particular, facial biometrics are utilized as primary identifiers in passport scenarios to facilitate secure border control and other identification verification applications. Facial biometrics can be captured live or through a passport photo submitted during the application process for identity-document issuance protocols. These biometrics are then stored in an identity document, such as an e-passport, which can be used for verification purposes as needed. Moreover, the implementation of e-passports will facilitate effortless Automatic Border Control (ABC) by eliminating the need for manual intervention through a comparison of the live image captured at the ABC gate with that of the electronic passport.

The accelerating adoption of the FRS technology is accompanied by an increase in vulnerability to various direct and indirect attacks. Among the several types of attacks on FRS, morphing attacks have gained prominence owing to their relevance in high-security applications, such as border control. Morphing involves seamless transformation of multiple face images into a single composite face image that exhibits the geometric and textural features of the original images. Morphing facial images have the ability to deceive both human observers (including experienced border guards) \cite{Rancha22_HumanObserver_Arxiv} and automatic FRS \cite{venkatesh2021facesurvey} posing a threat to ID verification and ABC scenarios.  In the application process for a passport, an attacker may employ a face morphing image to obtain a legitimate passport, which can subsequently be utilized to enter the country through ABC gates. These factors have motivated researchers to investigate both the generation and detection of face-morphing attacks \cite{venkatesh2021facesurvey}.

 The generation of face morphing for 2D images has been extensively explored using both handcrafted (landmarks) and deep learning techniques, such as Generative Adversarial Networks (GANs) and Diffusion Models. However, there has been less research on 3D face morphing due to the challenges of  3D facial key point registration between point clouds. Early work \cite{Singh_3DFaceMorphing_TBIOM23} in this area addressed the problem by converting 3D face point cloud into 2D RGB images and depth maps and using landmark-based morphing to generate morphing 2D RGB image and depth map which are back projected for generating morphing face point cloud. The process of generating a 3D face morphing point cloud using point clouds can be described as follows: Given two facial point clouds from two distinct individuals, the objective is to create a facial morphing point cloud that has an average 3D coordinate and color for the corresponding points. To the best of our knowledge, there is no existing work on directly generating 3D face morphing using 3D point clouds. Therefore, this work aims to address this gap by exploring 3D face morphing generation using 3D point clouds.

In the realm of 3D face generation, the critical component is the reliable registration of point clouds, which enables the generation of high-quality 3D morphing that can effectively deceive 3D FRS and achieve maximum attack potential.  Although 3D point cloud registration has been extensively studied in the literature on object detection and classification, the challenge of registering non-rigid objects, such as human faces, remains. This is due to the lack of known 3D correspondences between the two human face point clouds and the need to estimate affine transformations for each sub-region of the face. Various point set registration techniques have been proposed, including Reducing Kernel Hilbert Space (RKHS), spline functions, Thin Plate Spline (TPS~\cite{chui2003new}), correlation-based~\cite{tsin2004correlation}, Gaussian Mixture Models (GMM~\cite{jian2010robust}), Coherent Point Drift (CPD~\cite{Myronenko10_CPD_TPAMI}), and Bayesian Coherent Point Drift (BCPD~\cite{BCPDReference}). In this work, we employed BCPD because of its ability to register non-rigid objects such as human faces accurately, robustness against target rotation, and the use of non-Gaussian kernels, which results in greater efficiency than other existing methods, and the algorithm guarantees convergence. The following are the main contributions of this work: 

\begin{itemize} [leftmargin=*,noitemsep, topsep=0pt,parsep=0pt,partopsep=0pt]
    \item First work on generating 3D face morphing utilizing point clouds, leveraging the Bayesian Coherent Point Drift (BCPD) method for alignment and averaging the 3D coordinates and color from the given point clouds. 
    \item Extensive analysis on the publicly available 3D face dataset Facescape~\cite{Yang_2020_Facescape} with 200 unique identities. The attack potential of the proposed 3D morphing generation is evaluated using five different 3D FRS and two different 2D FRS.  
     \item The quantitative values of the attack potential is evaluated is Generalised Morphing Attack Potential (G-MAP) metric and compared with the existing 3D morphing generation techniques.
     \item The dataset and source code is available for research purpose. Link will be added in the final version. 
\end{itemize}

In the rest of the paper, we present the proposed method in Section~\ref{Sec:proposed} followed by Dataset details in Section~\ref{Sec:dataset}, experiments and results in Section~\ref{Sec:results} and  Section~\ref{Sec:conclusions} discuss the conclusion. 

\section{Proposed Method}\label{Sec:proposed}
\begin{figure*}
\centering
\includegraphics[width=0.8\textwidth]{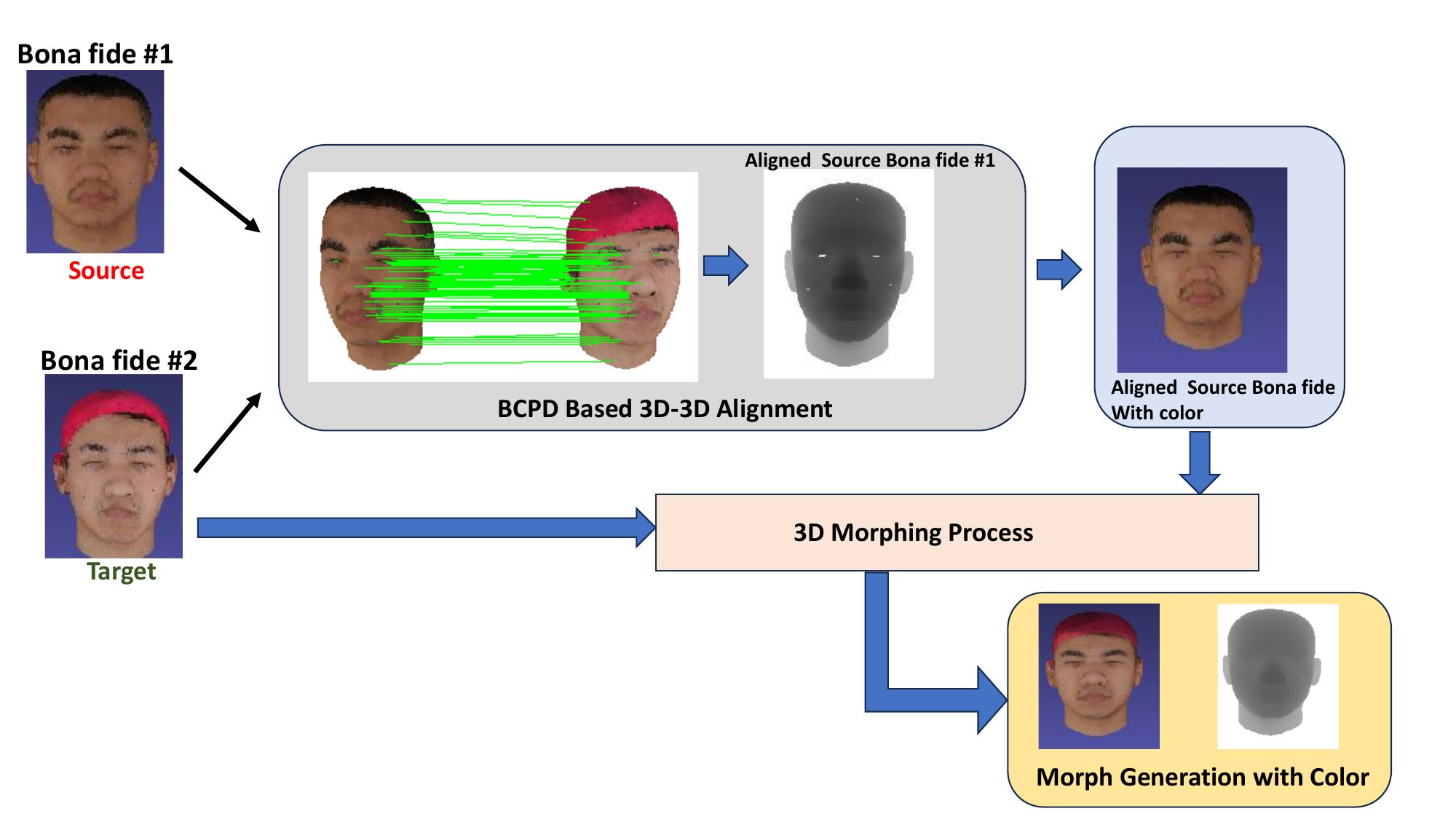}
\caption{Illustration showing block diagram of the proposed approach}
\label{fig:blockDiagram}
\end{figure*}%
\begin{figure*}
\centering
\includegraphics[width=0.7\textwidth]{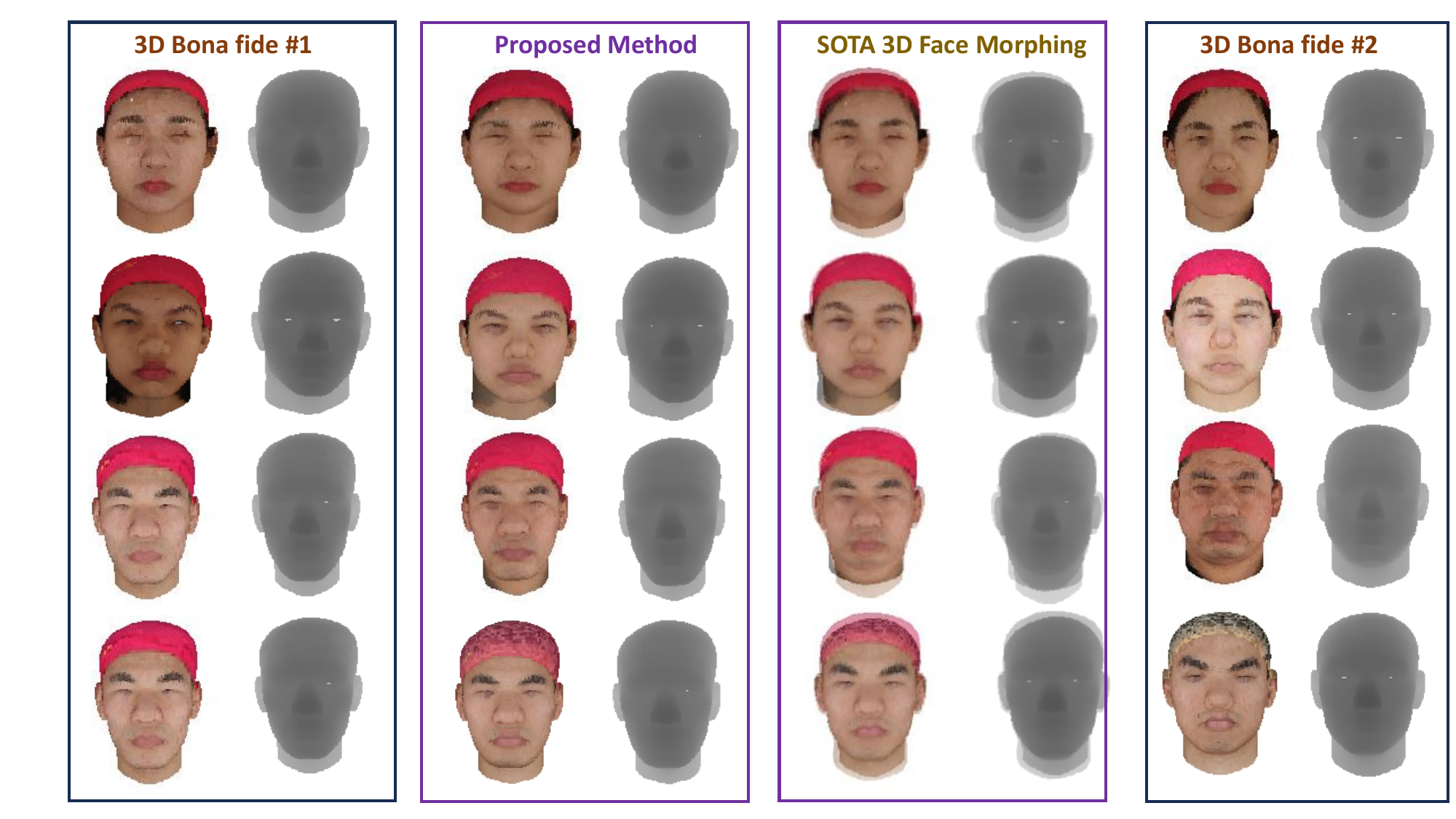}
\caption{Illustration showing Bona fide Input and Morphing Face Samples generated using proposed method and SOTA~\cite{Singh_3DFaceMorphing_TBIOM23} where SOTA shows blending artifacts in facial boundaries.}
\label{fig:datasetSamples}
\end{figure*}%


Figure \ref{fig:blockDiagram} shows the block diagram of the proposed method for 3D face morphing generation that can be structured in three different steps: (a) 3D point clouds alignment using Bayesian Coherent Point Drift (BCPD~\cite{BCPDReference}) (b) Colorization of the aligned point clouds (c) 3D morphing point clouds generation. Given two bona fide point clouds that are to be morphed, the proposed method will generate the 3D morphing cloud points as discussed below. 

\subsection{BCPD-based 3D-3D Alignment}
We adapted the BPCD algorithm \cite{BCPDReference} to perform 3D point cloud registration corresponding to two bona fide subjects. We first present the notations that are used to present the adapted BPCD algorithm, and then present the different steps of the 3D point cloud registration. 
\subsubsection{Notation}
\begin{itemize}
    \item Let $Ps_1,Ps_2$ denote the two input point clouds where ($Ps_1$) is considered as \textit{source point cloud} and ($Ps_2$) is considered as the \textit{target point cloud}.
    \item Let D denote the dimensionality of data, which in our case is $3$ because of 3D point clouds.
    \item Let the source point cloud be denoted as \\$Ps_1$=$(y_1^T,y_2^T,\cdots,y_M^T) \in \mathbb{R}^{3}$
    \item Let the target point cloud be denoted as\\ $Ps_2$=$(x_1^T,x_2^T,\cdots,x_N^T) \in \mathbb{R}^{3}$
    \item Let $Cs_1$ be the colors of source point cloud and $Cs_2$ be the colors of target point cloud.
    \item Let the displacement vectors obtained during the non-rigid\\ transformation be denoted as V=$(v_1^T,v_2^T,\cdots,v_M^T) \in \mathbb{R}^{3}$
    \item Let the similarity transform be denoted as $\rho=(s,R,T)$
    \item Let the multivariate normal distribution of $z$ with mean $\mu$ and co-variance matrix $S$ be denoted as $\phi(z;\mu,S)$
    \item Let the non-rigid transformation be denoted as\\ $T(y_M)=sR(y_M+v_M)+t$
    \item Let $G=g_{mm^\prime} \in \mathbb{R}^{M\mathrm{\times}M}$ be the Gram-Matrix with $g_{mm^\prime}=\kappa(y_M,y_M\prime)$ where $\kappa(.,.)$ is a positive-definite kernel.
    \item Let $P=(p_{mn}) \in [0,1]^{M\mathrm{\times}N}$ be the probability matrix where $p_{mn}$ represents the posterior probability that $x_n$ corresponds to $y_m$.
    \item Let $\nu=(\nu_1,\nu_2,\cdots,\nu_m)$ denote the estimated number of target points matched with each source points i.e. ($\nu_m=\sum_{n=1}^{N}p_{mn}$) 
    \item Let $Pst_1$ denote the transformed source point cloud.
    \end{itemize}
\subsubsection{Initialization}

The process of optimizing  typically begins with the initialization of various variables. In our case, we followed the BCPD algorithm~\cite{BCPDReference} to initialize the main variables, which include the Rotation Matrix ($R$), the Translation Vector ($t$), and $\sigma^2$. The Rotation Matrix was initialized to the identity matrix, the Translation Vector was initialized to zeros, and $\sigma^2$ was initialized based on the pairwise Euclidean distance between the source and target point clouds. The specific steps involved in this initialization process are outlined below.

\begin{itemize}
    \item $\hat{y}=y$,  $\hat{v}=0$, $\Sigma=I_M$, $s=1$, $R=I_D$, $t=0$, $\langle \alpha_M \rangle=\frac{1}{M}$ $\sigma^2=\frac{\gamma}{NMD}\sum_{n=1}^
    {N}\sum_{m=1}^
    {M}||x_n-y_m||^2$, $G=\langle g_{mm\prime} \rangle$ with $\langle g_{mm\prime} \rangle = \kappa(y_m,y_{m\prime})$
\end{itemize}
\subsubsection{Optimization}
Repeat the following steps until convergence.
\begin{itemize}
    \item Firstly, the probability matrix $P$ and its related variables are updated. The update is done based on existing variables. The variables updated in this step include $\langle \phi_{mn} \rangle$, $p_{mn}$, $\nu$, ${\nu\prime}$ and $\hat{x}$.
     \item Next update the following terms, the displacement variable ($\hat{v}$), covariance matrix ($\sum$) and related variables ($\Tilde{\nu},\Tilde{u}$ and $\lvert \alpha_m \rvert$ for all m) are updated in this step.
     \item Finally, update the parameters of transformation scaling (s), rotation matrix (R), translation vector (t) and the related variables ($\sigma^2$,$\hat{y}$).
\end{itemize}
Note the details about the updating of variables mentioned in previous step are available in~\cite{BCPDReference} (specifically in Figure 2 of the paper).
Once convergence is reached, the vertex-coordinates coordinates in the source point cloud can be transformed by the following equation:
\begin{equation}\label{eqn1}
Pst_1 = sR(Ps_1+V)+T
\end{equation}

\subsection{Colorization of the aligned point clouds}
The BCPD method does not transform per-vertex colors. Therefore, the transformation of Rotation, Displacement, and Translation is applied only to the point coordinates of the colored source point cloud generated by BCPD. The target point cloud is kept fixed, and the colors from the aligned point clouds are averaged to generate a face-morphing point cloud. To generate an aligned source point cloud, the coordinates of the source point cloud $Ps_1$ are transformed using Equation~\ref{eqn1}, resulting in a color-preserved aligned source point cloud $Pst_1$.
\subsection{3D Morphing Process}
The steps of generating the 3D face morphing point cloud ($P_m)$ given the aligned source point cloud ($Pst_1$) and the target point cloud ($Ps_2$) is done by generating the face morphing point cloud ($P_m$) vertices and colors by Equation~\ref{morph:vertex} and Equation~\ref{morph:color}, respectively. 
\begin{equation}\label{morph:vertex}
P_m=\alpha{\times}Pst_1+(1-\alpha){\times}Ps_2
\end{equation}
\begin{equation}\label{morph:color}
C_m=\alpha{\times}Cs_1+(1-\alpha){\times}Cs_2
\end{equation}

Figure~\ref{fig:datasetSamples} shows the qualitative results of the 3D face morphing generation using proposed and the state-of-the-art \cite{Singh_3DFaceMorphing_TBIOM23}. SOTA method generates facial morphing samples in which blending is noticeable at the boundaries whereas the proposed method blending is not visible. The blending is noticeable in depth-maps generated using SOTA apart from color images. The quality of the morphs generated by both methods appears to be similar in interior regions.



\section{Dataset Details}\label{Sec:dataset}
 In this work, we employed the publicly available 3D face dataset, Facescape~\cite{Yang_2020_Facescape}. Facescape dataset comprised 18,760 textured 3D faces from 938 data subjects captured with 20 different expressions. In this work, we selected 200 data subjects (112 males and 88 females) with neutral and smiling expressions to perform the morphing operation using both the proposed and existing 3D face morphing generation techniques. Morphing between the data subjects was performed by following the guidelines from \cite{frontex},  which include gender- and ethnicity-specific subject selection for morphing. The 3D face point clouds corresponding to neutral expressions were used to generate the morphing generation, and the smiling expression was used to compute the attack potential of the proposed morphing technique. 
The number of face morphing samples generated were 388 in total which includes 218 male morphs and 170 female morphs. 

\section{Experiments \& Results}\label{Sec:results}

In this section, we present a quantitative analysis of the attack potential of both the proposed and existing methods for 3D face morphing generation. This analysis was conducted using two different FRS: one based on depth maps using five pre-trained deep CNNs: Resnet-34, Inceptionv3, VGG16, Mobilenetv2 proposed in \cite{jiang2021pointface} and the the second FRS based on the depth maps, as proposed in \cite{Led3d2019}. Additionally, we utilized 2D FRS consisting of ArcFace and MagFace, which evaluates the attack potential of morphing attacks using color-morphed images without depth maps.

 In this work, Generalized Morphing Attack Potential (G-MAP) \cite{Singh23_CFIA_Access} which is a quantitative measure is  used to evaluate the attack potential of morphing images. G-MAP metric was designed to address the limitations of other evaluation metrics, as discussed in \cite{Singh23_CFIA_Access}.\footnote{A more detailed discussion about the advantages of G-MAP compared to other metrics is provided in \cite{Singh23_CFIA_Access}} To compute the G-MAP values, the morphing sample was enrolled in the FRS, and comparison scores were computed by probing the samples of the contributing subjects. If the computed scores exceed the False Acceptance Rate (FAR) threshold, the enrolled sample is considered a successful attack. Therefore, higher values of G-MAP correspond to a higher attack potential for morphing techniques. The G-MAP metric is defined as follows ~\cite{Singh23_CFIA_Access}:
 
 \begin{equation}\label{eqn:G-MAP}
\begin{aligned}
&{\textrm{G-MAP}} ={\frac{1}{|\mathbb{D}|}}{\sum_{d}^{|\mathbb{D}|}}{\frac{1}{|\mathbb{P}|}}{\frac{1}{|\mathbb{M}_d|}} \min_{\mathbb{F}_{l}}\\
& \sum_{i,j}^{|\mathbb{P}|,|\mathbb{M}_d|}\bigg\{\left[ (S1_i^j > \tau_l) \wedge \cdots ( Sk_i^j > \tau_l) \right]\\ &{\times}  \left[ (1-FTAR(i,l)) \right] \bigg\}\\
\end{aligned}
\end{equation}
where, $\mathbb{P}$ is set of probe images, $\mathbb{F}$ is the set of FRS, $\mathbb{D}$ is the set of Morphing Attack Generation Type, $\mathbb{M}_d$ is the face morphing image set for the Morphing Attack Generation Type $d$, $\tau_l$ indicate the similarity score threshold for FRS ($l$),  $FTAR(i,l)$ is the failure to acquire probe image in attempt $i$ using  FRS ($l$), and $||$ is the number of elements in a set.

In this work, we present the quantitative results for G-MAP with multiple probe attempts (G-MAP-MA) calculated from Equation \ref{eqn:G-MAP} by setting D = 1, F = 1, and FTAR = 0. We also present the G-MAP with multiple attempts and multiple FRS (G-MAP-MAMF) by taking the minimum across the FRS with  D = 1 in Equation \ref{eqn:G-MAP}. 

\begin{table}[htp]
\centering
\caption{G-MAP-MA @FAR = 0.1\%}
\resizebox{1.0\width}{!}{%
\begin{tabular}{|l|l|l|}
\hline
Algorithm/Features       & 3D Face Morphing~\cite{Singh_3DFaceMorphing_TBIOM23} & Proposed Method \\ \hline
\multicolumn{3}{|c|}{{\bf{3D FRS}}} \\ \hline
Resnet34~\cite{jiang2021pointface} & 86.79 & 97.93 \\ \hline
Inceptionv3~\cite{jiang2021pointface} & 81.61 & 97.93 \\ \hline
VGG16~\cite{jiang2021pointface} & 96.37 & 97.93 \\ \hline
Mobilenetv2~\cite{jiang2021pointface} & 97.67 & 97.93 \\ \hline
Led3D~\cite{Led3d2019} & 100 & 100 \\ \hline
\multicolumn{3}{|c|}{{\bf{2D FRS}}} \\ \hline
Arcface~\cite{deng2019arcface} & 95.85 & 100 \\ \hline
Magface~\cite{Meng21-Magface-CVPR} & 87.82 & 100 \\ \hline 
\end{tabular}%
}
\label{tab:GMAPTable1}%
\end{table}

 Table~\ref{tab:GMAPTable1} shows the Generalized Morphing Attack Potential (G-MAP) of multiple attempts using different face recognition systems (FRS) based on the depth maps. Based on the obtained results, it can be noted that (a) the proposed 3D face morphing generation techniques indicate higher values of GMAP, and thus indicate a higher attack potential compared to the existing method \cite{Singh_3DFaceMorphing_TBIOM23}. Improved performance of the proposed method was observed with both 3D and 2D FRS. The improved performance can be attributed to the high-quality color and depth maps, which can result in the vulnerability of the FRS. (b) With 3D FRS, the proposed method exhibited the best performance of GMAP = 97.93\% with deep CNNs and 100\% with Led3D FRS. With 2D FRS, the proposed method exhibits the best performance with GMAP = 100\% on both FRS.  
(c) Overall, there is a slight difference between SOTA and the proposed method, where SOTA shows blending artifacts in facial boundaries compared with the proposed method, which can also be seen in Figure~\ref{fig:datasetSamples}. This resulted in a lower G-MAP score with SOTA than with the proposed method. SOTA is based on a 3D-2D-3D approach, where blending is performed in 2D and can result in artifacts.

We also compute the GMAP-MAMF, which can quantify the attack potential of the generated morphing samples across multiple attempts and the FRS. For 3D FRS, the proposed morphing generation technique indicated a GMAP of 97.93\%, whereas the SOTA was 81.61\%.  For the 2D FRS, the proposed method indicates GMAP = 100\%, whereas SOTA is 87.82\%.  These results justify the higher attack potential of the proposed method compared to the existing method.

\section{Conclusions \& Future-Work}\label{Sec:conclusions}
In this paper, we introduced a method for directly registering 3D point clouds to generate a face-morphing point cloud based on BCPD. We evaluated the proposed 3D face morphing attack generation method on a publicly available dataset (Facescape Database) containing 200 unique data subjects. The attack potential of the proposed method was compared to that of the existing method using the G-MAP metric, and the results demonstrated the highest attack potential, as indicated by the quantitative analysis.
In the present work, the poses of the subjects were predominantly near-frontal, which simplifies the registration process. Moving forward, we plan to develop a method that can handle arbitrary facial positions and lighting conditions. This approach would be more representative of real-world scenarios because it would enable data capture under a variety of lighting conditions.


{\small
\bibliographystyle{ieee}
\bibliography{main}

\begin{thebibliography}{10}\itemsep=-1pt

\bibitem{chui2003new}
H.~Chui and A.~Rangarajan.
\newblock A new point matching algorithm for non-rigid registration.
\newblock {\em Computer Vision and Image Understanding}, 89(2-3):114--141, 2003.

\bibitem{deng2019arcface}
J.~Deng, J.~Guo, N.~Xue, and S.~Zafeiriou.
\newblock Arcface: Additive angular margin loss for deep face recognition.
\newblock In {\em Proceedings of the IEEE/CVF Conference on Computer Vision and Pattern Recognition}, pages 4690--4699, Long Beach, CA, USA, 2019.

\bibitem{frontex}
Frontex.
\newblock {\em Best Practice Technical Guidelines for Automated Border Control (ABC) Systems}.
\newblock Frontex, 2015.

\bibitem{Rancha22_HumanObserver_Arxiv}
S.~R. Godage, F.~Løvåsdal, S.~Venkatesh, K.~Raja, R.~Ramachandra, and C.~Busch.
\newblock Analyzing human observer ability in morphing attack detection -where do we stand?
\newblock {\em IEEE Transactions on Technology and Society}, pages 1--21, 2022.

\bibitem{BCPDReference}
O.~Hirose.
\newblock A bayesian formulation of coherent point drift.
\newblock {\em IEEE Transactions on Pattern Analysis and Machine Intelligence}, 43(7):2269--2286, 2021.

\bibitem{jian2010robust}
B.~Jian and B.~C. Vemuri.
\newblock Robust point set registration using gaussian mixture models.
\newblock {\em IEEE transactions on pattern analysis and machine intelligence}, 33(8):1633--1645, 2010.

\bibitem{jiang2021pointface}
C.~Jiang, S.~Lin, W.~Chen, F.~Liu, and L.~Shen.
\newblock Pointface: Point set based feature learning for 3d face recognition.
\newblock In {\em 2021 IEEE International Joint Conference on Biometrics (IJCB)}, pages 1--8, Shenzhen, China, 2021. IEEE.

\bibitem{Meng21-Magface-CVPR}
Q.~Meng, S.~Zhao, Z.~Huang, and F.~Zhou.
\newblock Magface: A universal representation for face recognition and quality assessment.
\newblock In {\em Proceedings of the IEEE/CVF conference on computer vision and pattern recognition}, pages 14225--14234, 2021.

\bibitem{Led3d2019}
G.~Mu, D.~Huang, G.~Hu, J.~Sun, and Y.~Wang.
\newblock Led3d: A lightweight and efficient deep approach to recognizing low-quality 3d faces.
\newblock In {\em Proceedings of the IEEE/CVF Conference on Computer Vision and Pattern Recognition (CVPR)}, June 2019.

\bibitem{Myronenko10_CPD_TPAMI}
A.~Myronenko and X.~Song.
\newblock Point set registration: Coherent point drift.
\newblock {\em IEEE transactions on pattern analysis and machine intelligence}, 32(12):2262--2275, 2010.

\bibitem{schroff2015facenet}
F.~Schroff, D.~Kalenichenko, and J.~Philbin.
\newblock Facenet: A unified embedding for face recognition and clustering.
\newblock In {\em 2015 IEEE Conference on Computer Vision and Pattern Recognition (CVPR)}, pages 815--823, Boston, MA, USA, 2015.

\bibitem{Singh_3DFaceMorphing_TBIOM23}
J.~M. Singh and R.~Ramachandra.
\newblock 3d face morphing attacks: Generation, vulnerability and detection.
\newblock {\em IEEE Transactions on Biometrics, Behavior, and Identity Science}, pages 1--1, 2023.

\bibitem{Singh23_CFIA_Access}
J.~M. Singh and R.~Ramachandra.
\newblock Deep composite face image attacks: Generation, vulnerability and detection.
\newblock {\em IEEE Access}, 11:76468--76485, 2023.

\bibitem{tsin2004correlation}
Y.~Tsin and T.~Kanade.
\newblock A correlation-based approach to robust point set registration.
\newblock In {\em Computer Vision-ECCV 2004: 8th European Conference on Computer Vision, Prague, Czech Republic, May 11-14, 2004. Proceedings, Part III 8}, pages 558--569. Springer, 2004.

\bibitem{venkatesh2021facesurvey}
S.~Venkatesh, R.~Ramachandra, K.~Raja, and C.~Busch.
\newblock Face morphing attack generation \& detection: A comprehensive survey.
\newblock {\em IEEE Transactions on Technology and Society}, 2021.

\bibitem{Yang_2020_Facescape}
H.~Yang, H.~Zhu, Y.~Wang, M.~Huang, Q.~Shen, R.~Yang, and X.~Cao.
\newblock Facescape: A large-scale high quality 3d face dataset and detailed riggable 3d face prediction.
\newblock In {\em IEEE/CVF Conference on Computer Vision and Pattern Recognition (CVPR)}, June 2020.

\end{thebibliography}
}

\end{document}


\ifFGfinal
\thispagestyle{empty}
\pagestyle{empty}
\else
\author{Anonymous FG2024 submission\\ Paper ID \FGPaperID \\}
\pagestyle{plain}
\fi
\maketitle


\begin{figure*}[h!]
        \centering
        \begin{subfigure}[b]{0.30\textwidth}
                \centering
                \includegraphics[width=\textwidth]{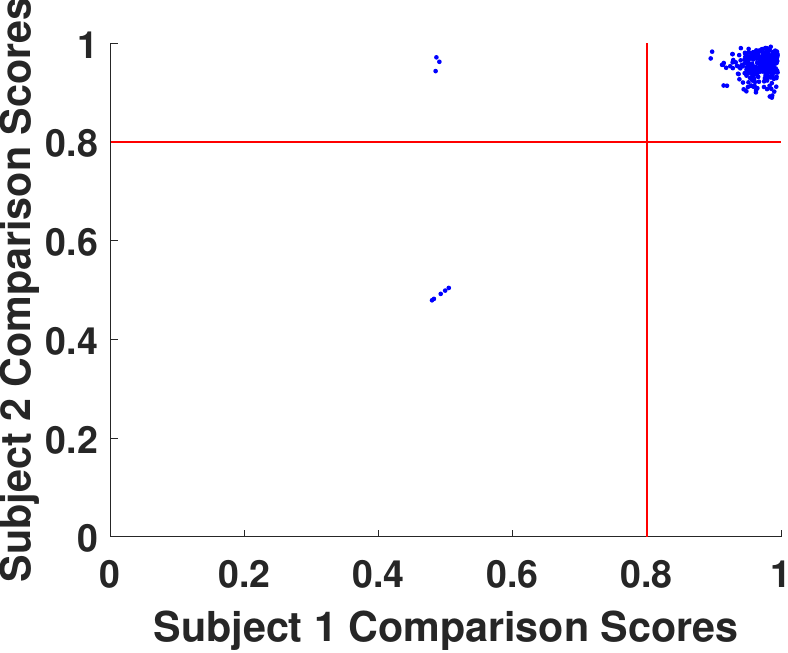}
                \caption{}
                \label{fig:gull}
        \end{subfigure}%
        \begin{subfigure}[b]{0.30\textwidth}
                \centering
                \includegraphics[width=\textwidth]{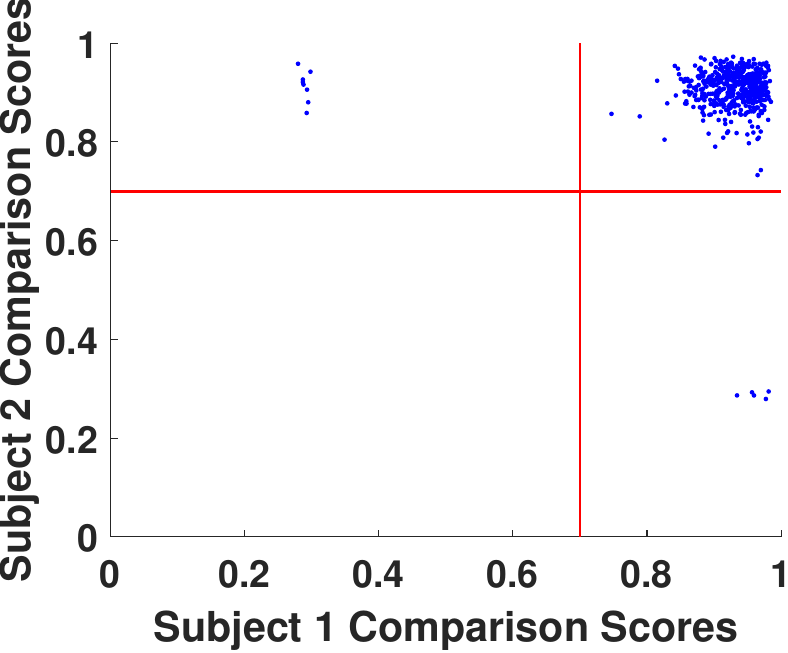}
                \caption{}
                \label{fig:gull2}
        \end{subfigure} 
         \begin{subfigure}[b]{0.30\textwidth}
                \centering
                \includegraphics[width=\textwidth]{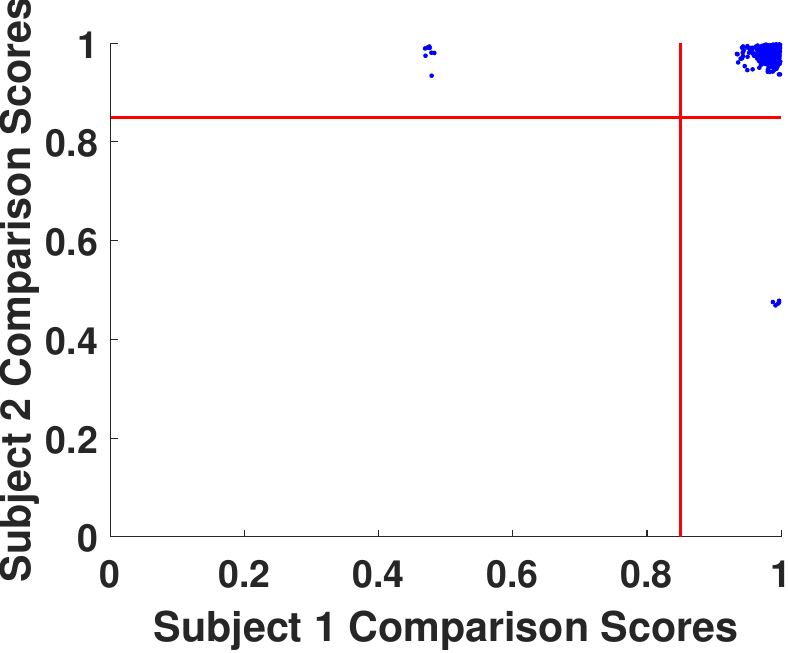}
                \caption{}
                \label{fig:gull2}
        \end{subfigure} \\ 
             \begin{subfigure}[b]{0.30\textwidth}
                \centering
                \includegraphics[width=\textwidth]{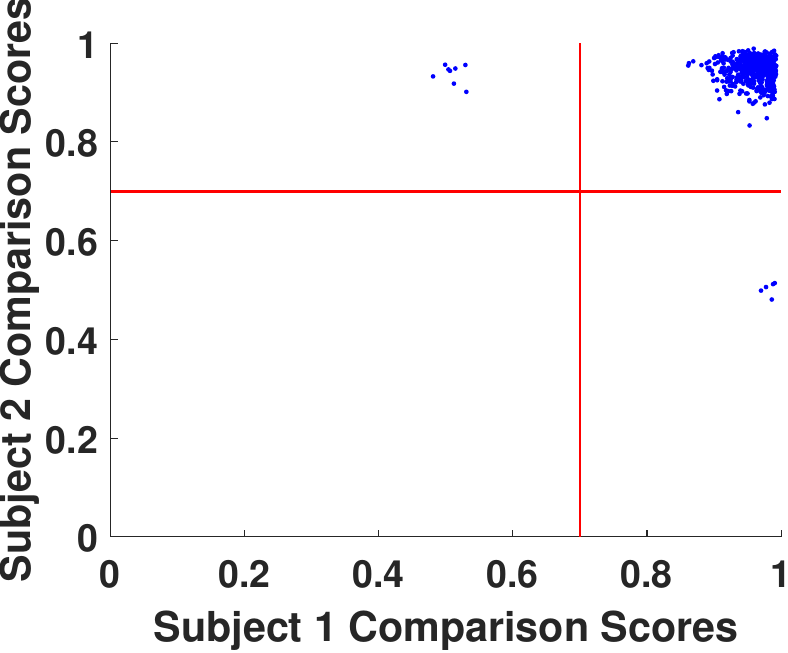}
                \caption{}
                \label{fig:gull2}
        \end{subfigure} 
        \begin{subfigure}[b]{0.30\textwidth}
                \centering
                \includegraphics[width=\textwidth]{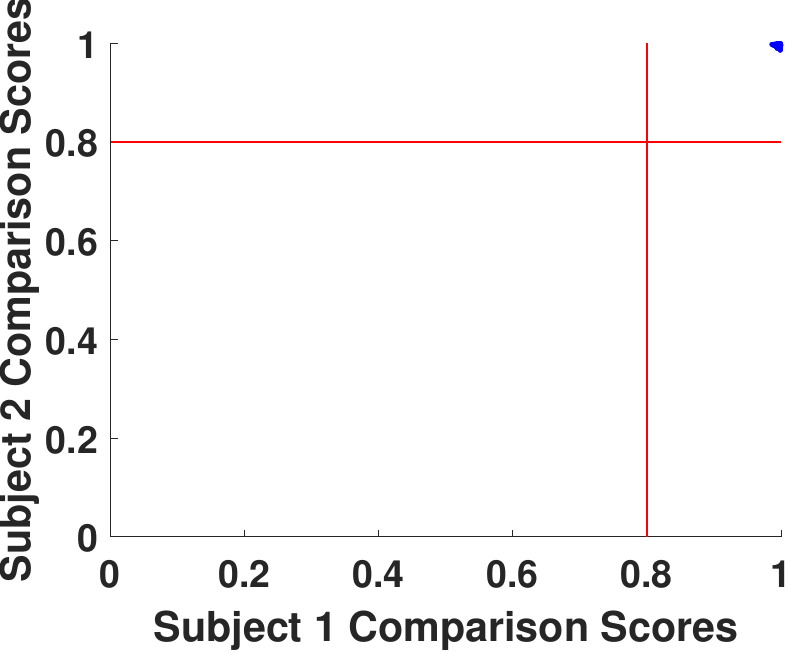}
                \caption{}
                \label{fig:gull2}
                \end{subfigure} 
        \begin{subfigure}[b]{0.30\textwidth}
                \centering
                \includegraphics[width=\textwidth]{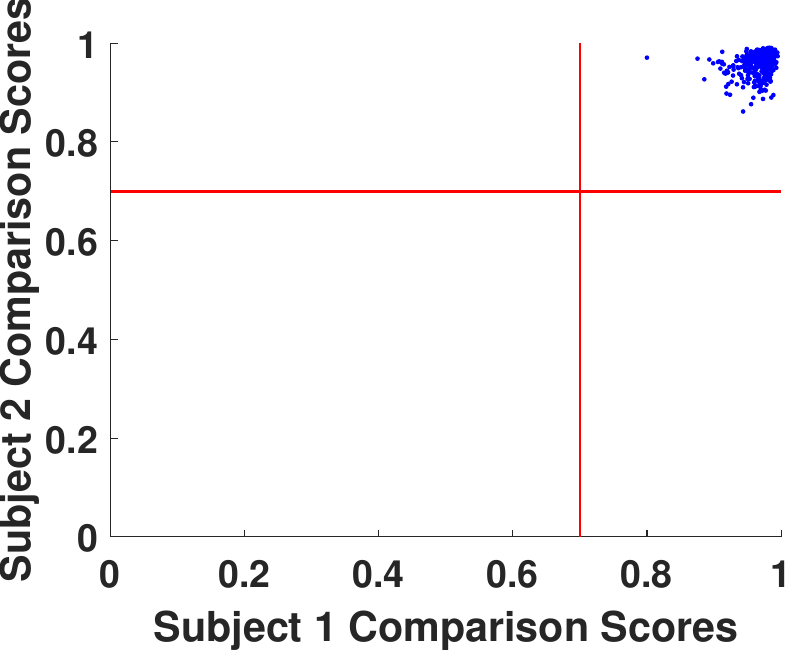}
                \caption{}
                \label{fig:gull2}
        \end{subfigure}\\
        \begin{subfigure}[b]{0.30\textwidth}
                \centering
                \includegraphics[width=\textwidth]{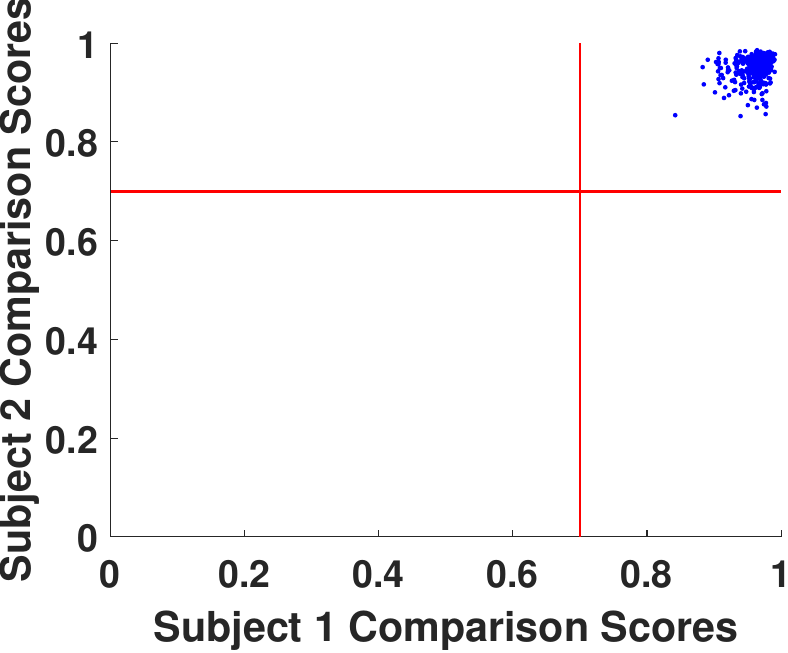}
                \caption{}
                \label{fig:gull2}
            \end{subfigure}
            
        \caption{{Vulnerability Plots using Proposed Method on Facescape Dataset  (a) Resnet34, (b) Inceptionv3, (c) VGG16, (d) Mobilenetv2, (e) Led3D~\cite{Led3d2019}, (f) Arcface~\cite{deng2019arcface} (g) Magface~\cite{Meng21-Magface-CVPR}}}\label{fig:Vulnerability3DFaceMorphing}
\end{figure*}

\begin{figure*}[h!]
        \centering
        \begin{subfigure}[b]{0.30\textwidth}
                \centering
                \includegraphics[width=\textwidth]{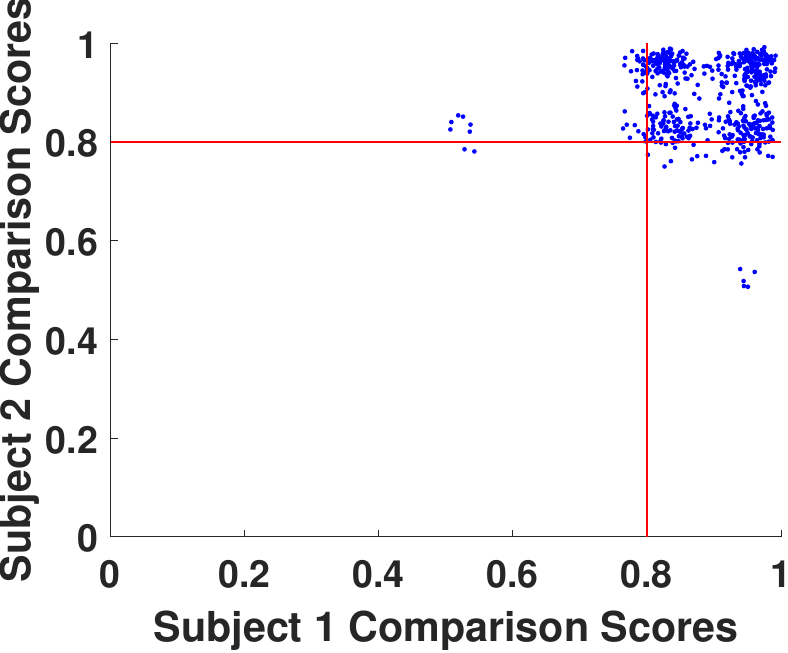}
                \caption{}
                \label{fig:gull}
        \end{subfigure}%
        \begin{subfigure}[b]{0.30\textwidth}
                \centering
                \includegraphics[width=\textwidth]{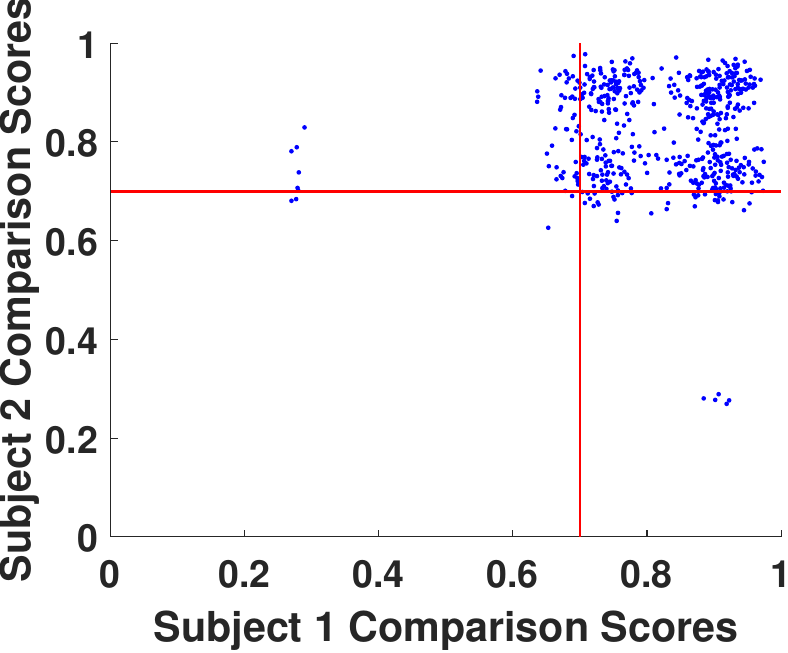}
                \caption{}
                \label{fig:gull2}
        \end{subfigure} 
         \begin{subfigure}[b]{0.30\textwidth}
                \centering
                \includegraphics[width=\textwidth]{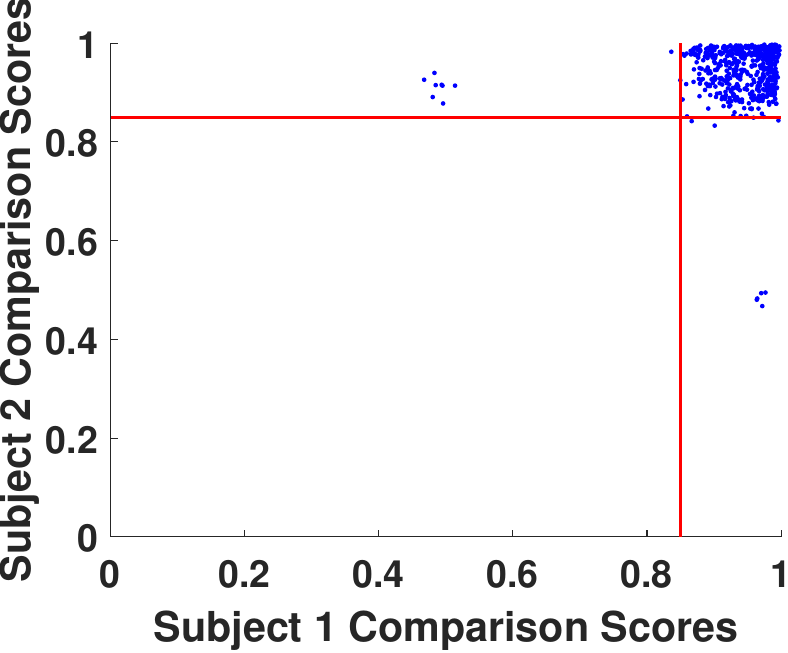}
                \caption{}
                \label{fig:gull2}
        \end{subfigure} \\ 
             \begin{subfigure}[b]{0.30\textwidth}
                \centering
                \includegraphics[width=\textwidth]{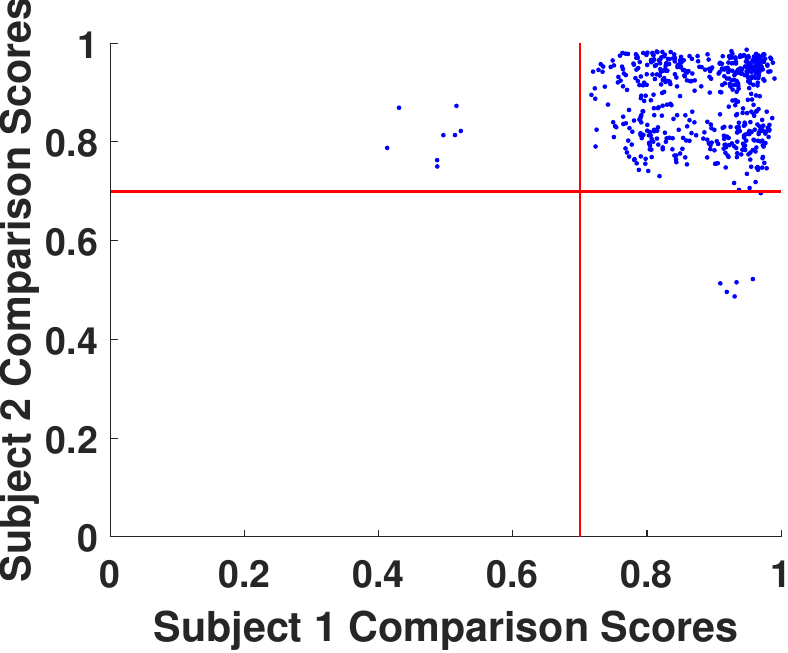}
                \caption{}
                \label{fig:gull2}
        \end{subfigure} 
        \begin{subfigure}[b]{0.30\textwidth}
                \centering
                \includegraphics[width=\textwidth]{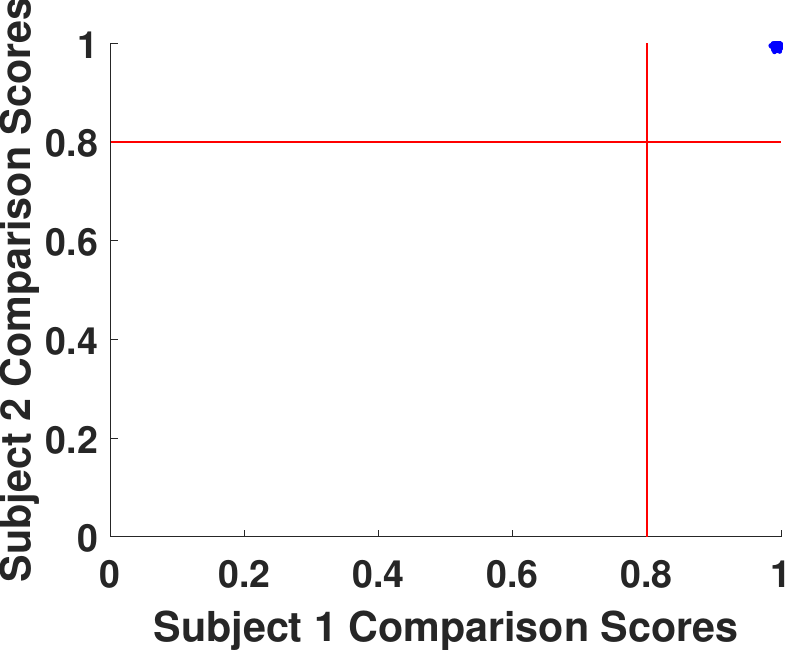}
                \caption{}
                \label{fig:gull2}
                
        \end{subfigure} 
        \begin{subfigure}[b]{0.30\textwidth}
                \centering
                \includegraphics[width=\textwidth]{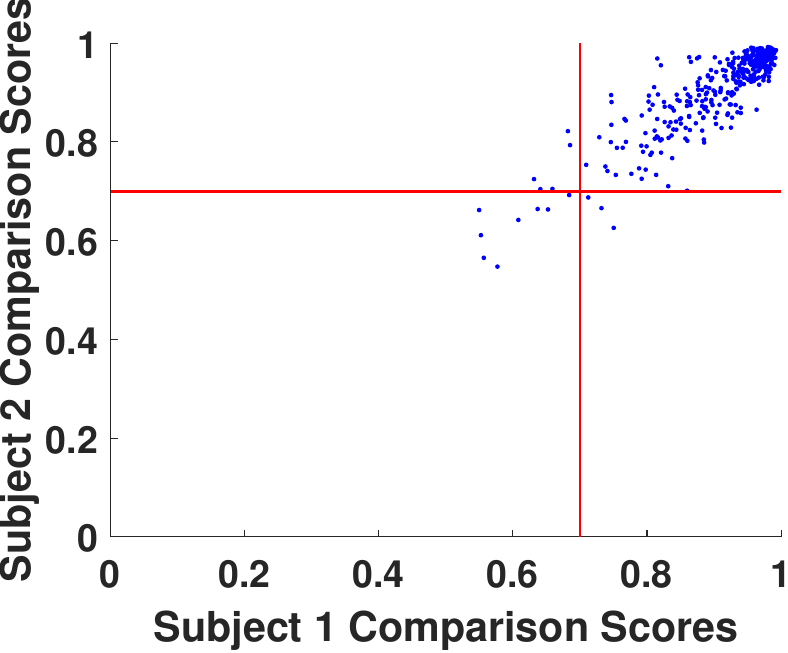}
                \caption{}
                \label{fig:gull2}
        \end{subfigure} \\
        \begin{subfigure}[b]{0.30\textwidth}
                \centering
                \includegraphics[width=\textwidth]{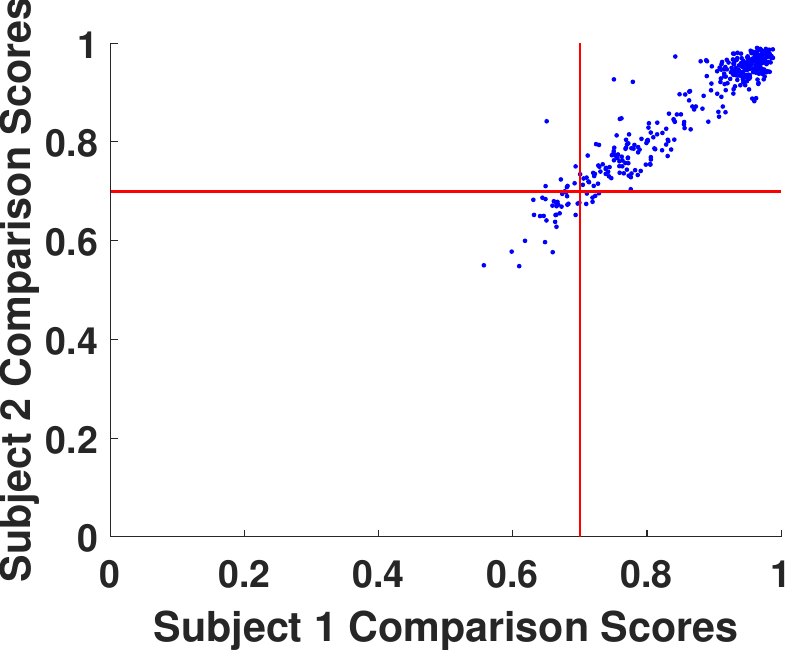}
                \caption{}
                \label{fig:gull2}
            \end{subfigure}
            
         \caption{{Vulnerability Plots using 3D Face Morphing~\cite{Singh_3DFaceMorphing_TBIOM23} on Facescape Dataset (a) Resnet34, (b) Inceptionv3, (c) VGG16, (d) Mobilenetv2 (e) Led3D~\cite{Led3d2019}, (f) Arcface~\cite{deng2019arcface} (g) Magface~\cite{Meng21-Magface-CVPR}}}\label{fig:Vulnerabilityproposed}
\end{figure*}

\section{Scatter Plots: Vulnerability Analysis}\label{supplementary}
 Figure~\ref{fig:Vulnerability3DFaceMorphing} and Figure~\ref{fig:Vulnerabilityproposed} presents the scatter plots of 2D and 3D FRS for SOTA~\cite{Singh_3DFaceMorphing_TBIOM23} and the proposed method, respectively. It needs to be pointed out that the face morphing sample ($M_S12$) is enrolled and probed against its two bona fide subjects, $S1$ and $S2$, resulting in comparison scores for Subject 1 and Subject 2, respectively. An enrolled face morphing sample is considered a threat only if both comparison scores exceed the threshold at False Match Rate=0.1\% (first (right-top) quadrant in the vulnerability score plot). Further, the scores in the second(left-top) and fourth(right-bottom) quadrants imply that one subject's scores are lower than the threshold. Finally, the third(left-bottom) quadrant means that the scores of both subjects are lower than the threshold. The proposed method (Figure~\ref{fig:Vulnerability3DFaceMorphing}) and SOTA(Figure~\ref{fig:Vulnerabilityproposed}) show a high percentage of scores exceeding the threshold, indicating high vulnerability. However, the proposed method indicates higher vulnerability than SOTA for the seven FRSs.

{\small
\bibliographystyle{ieee}
\bibliography{egbib}
}